\pdfoutput=1

\documentclass[11pt]{article}

\usepackage[final]{acl}

\usepackage{times}
\usepackage{latexsym}
\usepackage{enumitem}

\usepackage[T1]{fontenc}

\usepackage[utf8]{inputenc}

\usepackage{microtype}

\usepackage{inconsolata}

\usepackage{booktabs}
\usepackage{amssymb}
\usepackage{amsmath}
\usepackage{multirow}
\usepackage{bbm}
\usepackage{subcaption}
\usepackage{graphicx}
\title{Editing the Mind of Giants: An In-Depth Exploration of Pitfalls of Knowledge Editing in Large Language Models}

\author{Cheng-Hsun Hsueh$^*$\quad Paul Kuo-Ming Huang$^*$\quad Tzu-Han Lin$^*$\quad Che-Wei Liao$^*$ \\
\textbf{Hung-Chieh Fang$^*$\quad Chao-Wei Huang\quad Yun-Nung Chen}
\\National Taiwan University, Taipei, Taiwan \\
\texttt{\{r12922059,b08902072,r12944034,r09922a25\}@csie.ntu.edu.tw}\\
 \texttt{\{b09902106,f07922069\}@csie.ntu.edu.tw\quad y.v.chen@ieee.org}
}

\begin{document}
\maketitle
\begin{abstract}
Knowledge editing is a rising technique for efficiently updating factual knowledge in large language models (LLMs) with minimal alteration of parameters. However, recent studies have identified side effects, such as knowledge distortion and the deterioration of general abilities, that have emerged after editing. Despite these findings, evaluating the pitfalls of knowledge editing often relies on inconsistent metrics and benchmarks, lacking a uniform standard. In response, this survey presents a comprehensive study of these side effects, providing a unified perspective on the challenges of knowledge editing in LLMs by conducting experiments with consistent metrics and benchmarks. Additionally, we review related works and outline potential research directions to address these limitations. Our survey highlights the limitations of current knowledge editing methods, emphasizing the need for a deeper understanding of the inner knowledge structures of LLMs and improved knowledge editing methods.
To foster future research, we have released the complementary materials publicly\footnote{\url{https://github.com/MiuLab/EditLLM-Survey}}.
\begingroup\def\thefootnote{\rm *}\footnotetext{Equal contribution.}\endgroup
\end{abstract}

\section{Introduction}
Recent advancements in large language models (LLMs) have significantly improved NLP applications, enabling LLMs to understand and generate language at a human-like level. However, the mechanisms of knowledge storage in LLMs remain unclear, raising concerns about the reliability of their output, particularly in applications like chatbots. To address these issues, researchers have explored various methods. Traditional methods like fine-tuning, continual learning, and retraining are computationally expensive and may degrade LLM performance. \emph{Knowledge editing} has emerged as a promising alternative, offering efficient adjustments with minimal computational costs and fewer alterations~\citep{DeCao2021EditingFK,dai-etal-2022-knowledge,Meng2022LocatingAE,meng2023memit,dong-etal-2022-calibrating,mitchell2022fast,mitchell2022memory,hartvigsen2023aging,huang2023transformerpatcher,yu2023melo,zheng2023can,Li2023PMETPM,tan23malmen,gupta2024unified,Wang2024DeepEditKE}. This method allows precise LLMs refinement, enhancing their practical and reliable use in real-world applications.

\begin{figure}[t!]
\centering    \includegraphics[width=\linewidth]{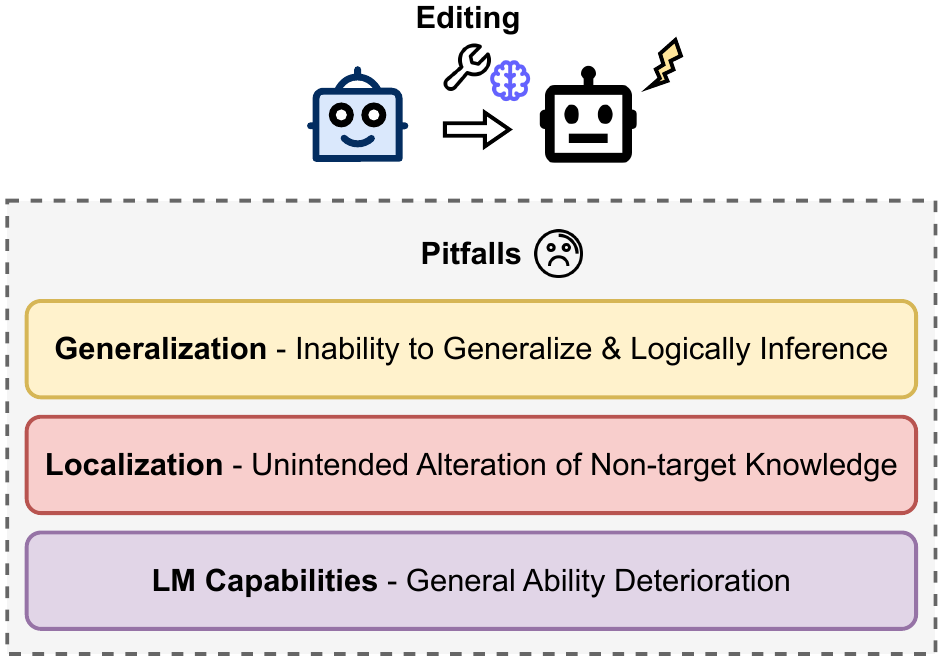}
    \caption{An overview of pitfalls in current knowledge editing methods. The subsequent sections dive into three key challenges: generalization issues (Section \ref{sec:prob-generalization}), locality issues (Section \ref{sec:prob-alteration}), and deterioration of general LLM abilities (Section \ref{sec:catastrophic}).}
    \label{fig:overview}
\end{figure}

Knowledge editing can be divided into two main categories: parameter-modifying and parameter-preserving. Both aim to refine LLM knowledge efficiently while avoiding the drawbacks of previous tuning methods~\citep{Yao2023EditingLL}. Parameter-modifying methods, including meta-learning~\citep{DeCao2021EditingFK,mitchell2022fast,tan23malmen} and locate-and-edit techniques~\citep{dai-etal-2022-knowledge,Meng2022LocatingAE,meng2023memit,Li2023PMETPM,gupta2024unified}, strive to update model parameters effectively. By contrast, parameter-preserving methods introduce external components, like knowledge bases~\citep{mitchell2022memory,zhong2023mquake} or extra model parameters~\citep{dong-etal-2022-calibrating,huang2023transformerpatcher,hartvigsen2023aging,yu2023melo} to maintain the integrity of pre-trained LLMs while updating their knowledge.

Despite the success of knowledge editing, challenges remain. Recent studies have revealed side effects that can harm the general capabilities and intrinsic structures of LLMs. We categorize these pitfalls into three main areas: (1) the inability to perform logical inference~\citep{Cohen2023EvaluatingTR, li2024unveiling, zhong2023mquake, hua2024propagation, Yao2023EditingLL}, (2) the unintended modification of non-target knowledge~\citep{Cohen2023EvaluatingTR, li2024unveiling, Yao2023EditingLL, Meng2022LocatingAE, hoelscher-obermaier-etal-2023-detecting}, and (3) the deterioration of general LLM abilities~\citep{gupta2024model, gu2024model, yang2024butterfly}. Although various side effects have been identified, they are evaluated using inconsistent metrics and benchmarks in different studies, which lack a uniform standard. As a result, this survey aims to provide a comprehensive overview of the current issues in the knowledge editing paradigm and to establish a fair platform for comparing the side effects of different editing methods. Additionally, we encourage further investigation into the pitfalls and underlying knowledge structures of LLMs. A brief overview of the discussed pitfalls is shown in Figure ~\ref{fig:overview}.

This paper is organized as follows: Section~\ref{sec:overview_edit} introduces the definition and methods of knowledge editing. Section~\ref{sec:challenges} discusses current challenges and corresponding benchmarks. In Section~\ref{sec:experiments}, we present experimental results evaluating different editing methods. Finally, Section~\ref{sec:limit_chall_future} explores related studies and future research directions. We summarize our contributions as follows:
\begin{enumerate}[nosep]
    \item We are the first to provide a comprehensive analysis of the side effects associated with existing knowledge editing techniques.
    \item We systematically organize previous research and conduct experiments to benchmark the side effects of knowledge editing, providing a unified perspective on this issue.
    \item We discuss related studies and potential directions to address existing challenges, encouraging further exploration in this field.
\end{enumerate}

\begin{figure}[t]
\centering
    \includegraphics[width=\linewidth]{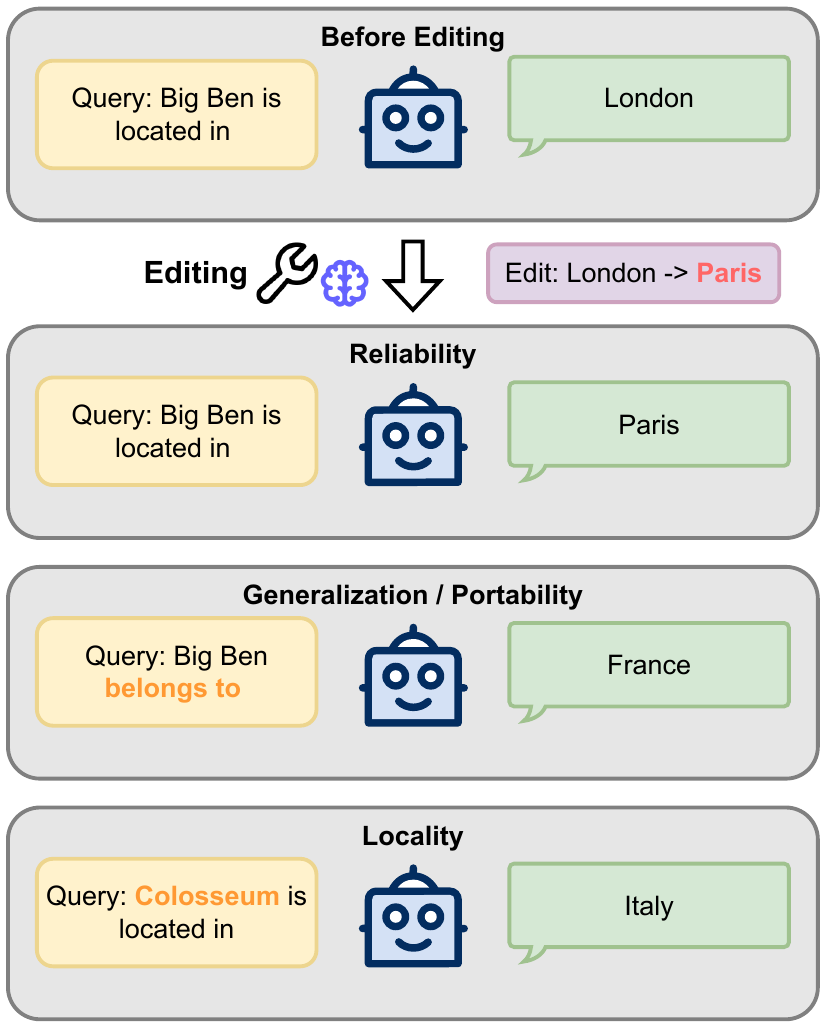}
    \caption{Illustration of properties that knowledge editing methods should satisfy: reliability, generalizability/portability, and locality.}
    \label{fig:basic_properties}
\end{figure}

\begin{figure*}[ht]
\centering
    \begin{subfigure}{0.75\linewidth}
    \includegraphics[width=\linewidth]{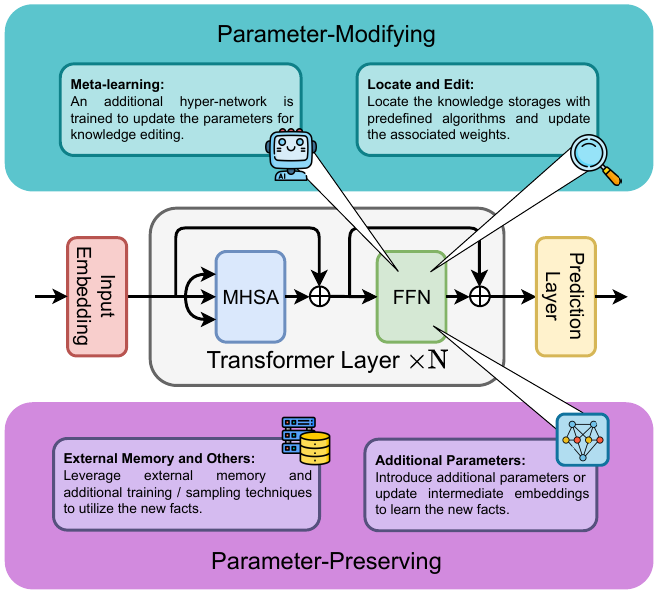}
    \end{subfigure}
    \caption{Illustration of the two categories of model editing methods in transformer-based large language models, which includes parameter-modifying (meta-learning and locate-and-edit) and parameter-preserving (additional parameters, external memory, in-context learning, and decoding) methods. MHSA and FFN stand for multi-head self-attention and feed-forward network, respectively.}
    \label{fig:edit_methods}
\end{figure*}

\section{Overview of Knowledge Editing}
\label{sec:overview_edit}

\subsection{Problem Definition}
Knowledge editing for LLMs entails modifying the output of LLMs in response to specific edit queries, with the aim of minimizing alterations to their original behavior~\citep{Yao2023EditingLL,Mazzia2023ASO,zhang2024comprehensive}. In this section, we follow the notation from~\citet{Mazzia2023ASO}.

We denote the input and output space as $\mathbb{X}$ and $\mathbb{Y}$, respectively. The function space $\mathbb{F}: \mathbb{X} \rightarrow \mathbb{Y}$ is estimated by the base model $f_{\theta_0}$ parameterized by $\theta_0 \in \Theta$. Finally, let $Z_e = \{ (x_e, y_e) \ | \ f_{\theta_0}(x_e) \neq y_e \}$ be the set of edit queries we would like to apply to the base model.
The goal of knowledge editing is to efficiently derive the edited model $f_{\theta_e}$ from the base model that satisfies the following:
\begin{equation}
    f_{\theta_e}(x_e) = y_e, \forall (x_e, y_e) \in Z_e
\end{equation}

The ideal edited model $f_{\theta_e}$ should satisfy three properties: \textbf{reliability}, \textbf{generalization}, and \textbf{locality}. An illustration is shown in Figure~\ref{fig:basic_properties}.

\paragraph{Reliability}
Given an edit query $(x_e, y_e)$, the edited model $f_{\theta_e}$ should output the target answer $y_e$ when given the target input $x_e$, i.e. $f_{\theta_e}(x_e) = y_e$. The reliability of a editing method is measured by calculating the average edit success rate:
\begin{equation}
    \mathbb{E}_{(x_e', y_e')\sim Z_e} \mathbbm{1}\{ f_{\theta_e}(x_e') = y_e' \}
\end{equation}

\paragraph{Generalization}
The edited model should generalize the edited knowledge to relevant instances. The generalization metric is commonly formulated as the average success rate on the neighboring set:
\begin{equation}
    \mathbb{E}_{(x_e', y_e')\sim N(x_e, y_e)} \mathbbm{1} \{ f_{\theta_e}(x_e') = y_e' \},
\end{equation}
where $N(x_e, y_e)$ is the set of neighboring instances of an edit query $(x_e, y_e)$. Earlier works evaluate this metric by rephrasing the input prompts~\citep{mitchell2022fast, Meng2022LocatingAE, huang2023transformerpatcher}.

\label{para:gen}
\paragraph{Locality}
The editing process should not affect instances unrelated to the edit queries. The locality set of an edit query $(x_e, y_e)$ can be defined as $L(x_e) = \{ (x_{loc}, y_{loc}) \in \mathbb{X} \times \mathbb{Y}\ \mathrm{s.t}\ x_{loc} \notin N(x_e, y_e) \land f_{\theta_0}(x_{loc}) = y_{loc} \}$. The locality, also known as specificity, of an editing method is measured by calculating the level of invariance of model output before and after the edits, which can be calculated as follows:
\begin{equation}
    \mathbb{E}_{(x_{loc}, y_{loc})\sim L(x_e)} \mathbbm{1} \{ f_{\theta_e}(x_{loc}) = y_{loc} \}
\end{equation}

\subsection{Current Methods}
Current knowledge editing methods are categorized into parameter-modifying (Section~\ref{subsec:parameters-modifying}) and parameter-preserving (Section~\ref{subsec:parameter-preserving}) editing methods, each containing several strategies. An overview and illustration of current methods are included in Table~\ref{tab:methods_v2} and Figure~\ref{fig:edit_methods}, respectively.

\begin{table}[t]
    \centering
    \scriptsize
    \begin{tabular}{ccl}
    \toprule
        Category & Strategy & Method \\
        \midrule
        \multirow{ 8}[3]{*}{\shortstack{Parameter- \\ modifying}}
            & \multirow{ 3}{*}{\shortstack{Meta- \\ learning}}
                & Knowledge Editor~\citep{DeCao2021EditingFK} \\
                & & MEND~\citep{mitchell2022fast} \\
                & & MALMEN~\citep{tan23malmen} \\
        \cmidrule{2-3}
            & \multirow{ 5}{*}{\shortstack{Locating \\ and \\ editing}}
                & Knowledge Neuron~\citep{dai-etal-2022-knowledge} \\
                & & ROME~\citep{Meng2022LocatingAE} \\
                & & MEMIT~\citep{meng2023memit} \\
                & & PMET~\citep{Li2023PMETPM} \\
                & & EMMET~\citep{gupta2024unified} \\
        \midrule
        \multirow{ 10}[3]{*}{\shortstack{Parameter- \\ preserving}}
            & \multirow{ 4}{*}{\shortstack{Additional \\ parameters}}
                & CaliNET~\citep{dong-etal-2022-calibrating} \\
                & & T-Patcher$^\dagger$~\citep{huang2023transformerpatcher} \\ 
                & & GRACE$^\dagger$~\citep{hartvigsen2023aging} \\ 
                & & MELO$^\dagger$~\citep{yu2023melo} \\
        \cmidrule{2-3}
            & \multirow{ 2}{*}{\shortstack{External \\ memory}}
                & SERAC$^\dagger$~\citep{mitchell2022memory}  \\
                & & MeLLo$^\dagger$~\citep{zhong2023mquake}  \\
        \cmidrule{2-3}
            & \multirow{ 2}{*}{\shortstack{In-context \\ learning}}
                & \multirow{ 2}{*}{IKE$^\dagger$~\citep{zheng2023can}} \\\\
        \cmidrule{2-3}
            & Decoding
                & DeepEdit$^\dagger$~\citep{Wang2024DeepEditKE} \\
    \bottomrule
    \end{tabular}
    \caption{Overview of knowledge editing methods. The methods are categorized into two major families, parameter-modifying and parameter-preserving methods, each containing several strategies. Methods marked with $\dagger$ have the ability to process sequential edits.}
    \label{tab:methods_v2}
\end{table}

\subsubsection{Parameter-Modifying}
\label{subsec:parameters-modifying}

\paragraph{Meta-learning}
Meta-learning methods train a hyper-network to predict network parameter updates. For instance, KnowledgeEditor~\citep{DeCao2021EditingFK} trains a deep network to predict weight updates. MEND~\citep{mitchell2022fast} decomposes the gradient matrix into two rank-one matrices and utilized a hyper-network to update these matrices, thereby accelerating the editing process. Built upon MEND, MALMEN~\citep{tan23malmen} refines the process by formulating the aggregation of parameter shifts into a least-squares problem, further improving the scalability of meta-learning methods.

\paragraph{Locate and Edit}
Locate-and-edit methods identify specific knowledge locations in LLMs for consequent editing. KN~\citep{dai-etal-2022-knowledge} utilizes the proposed knowledge attribution method to pinpoint neurons expressing relational facts, allowing efficient updates or erasures without fine-tuning. ROME~\citep{Meng2022LocatingAE} proposes causal tracing method to identify neuron activations linked to specific knowledge. The authors demonstrate the significance of middle-layer feed-forward networks (FFNs) in factual predictions when processing the subject's last token. Built upon the hypothesis that the FFN modules in a transformer layer can be viewed as key-value memories~\citep{geva-etal-2021-transformer}, ROME injects new knowledge into the key-value memories by deriving the closed form solution from the least-squares problem. MEMIT~\citep{meng2023memit} scales up ROME by editing a set of MLPs from consecutive middle-layers via solving a normal equation. PMET~\citep{Li2023PMETPM} proposes to update multi-head self-attention (MHSA) modules in addition to FFNs. EMMET~\citep{gupta2024unified} on the other hand, integrates the objectives of ROME and MEMIT into a unified preservation-memorization objective, facilitating batch-editing capabilities for both methodologies.

\subsubsection{Parameter-Preserving}
\label{subsec:parameter-preserving}


\paragraph{Additional Parameters} Some methods utilize additional parameters, such as adding new neurons or employing parameter-efficient techniques. CaliNET~\citep{dong-etal-2022-calibrating} extends the FFN modules with calibration memory slots to adjust the predicted token distribution. T-Patcher~\citep{huang2023transformerpatcher} adds neurons in the FFN’s last layer to rectify classification errors and incorrectly generated tokens, activating only in response to associated mistakes. GRACE~\citep{hartvigsen2023aging} wraps a selected layer with an Adaptor that includes a codebook and deferral mechanism, learning to decode desired outputs while caching embeddings of error inputs. The GRACE layer stores the edits and could be updated continuously over long deployments. MELO~\citep{yu2023melo} utilizes DyLoRA~\citep{valipour-etal-2023-dylora} modules to learn edits, indexing them in an inner vector database to dynamically activate corresponding LoRA blocks during inference.

\paragraph{External Memory} Other methods utilize external memories for editing. SERAC~\citep{mitchell2022memory} leverages a scope classifier to determine whether an user-supplied edit example stored in its memory is related to the inputs. If no example exists, the inputs are passed to the base model; otherwise, a counterfactual model generates modified answers using the inputs and the related example.  MeLLo~\citep{zhong2023mquake} decomposes a multi-hop question into subquestions iteratively. The model then checks if the tentative answer generated by the base model contradicts the most relevant facts retrieved from the edited fact memory and adjusts the outputs accordingly.

\paragraph{In-Context Learning and Decoding} Certain strategies require no additional parameters. IKE~\citep{zheng2023can} edits factual knowledge via in-context learning with demonstrations to guide the language model. DeepEdit~\citep{Wang2024DeepEditKE} employs decoding constraints, including filtering step candidates, depth-first search to store valid candidates in a stack, and a greedy search to output the optimal path for multi-hop reasoning.

\section{Challenges of Knowledge Editing}
\label{sec:challenges}
While knowledge editing methods have been extensively researched, comprehensive studies on related challenges are lacking. In this section, we discuss the pitfalls of knowledge editing from three perspectives: inability to logically infer and robustly generalize (Section~\ref{sec:prob-generalization}), unintended alteration of non-target knowledge (Section~\ref{sec:prob-alteration}), and deterioration of general LLM abilities (Section~\ref{sec:catastrophic}).

\begin{table*}[ht]
    \centering
    \footnotesize
    \begin{tabular}{clp{6.5cm}}
    \toprule
        Challenge & Benchmark & Metric  \\
        \midrule
        \multirow{ 7}[3]{*}{\shortstack{Portability \\ and \\ Generalization}}
            & \multirow{2}{*}{RippleEdits~\citep{Cohen2023EvaluatingTR}}
                & Logical Generalization, Compositionality I, Compositionality II \\
        \cmidrule(l){2-3}
            & \multirow{1}{*}{ConflictEdit~\citep{li2024unveiling}}
                & Conflict Score, Conflict Magnitude, Success Score \\
        \cmidrule(l){2-3}
            & \multirow{2}{*}{MQuAKE~\citep{zhong2023mquake}}
                & Edit-wise Success Rate, Instance-wise Accuracy, Multi-hop Accuracy\\
        \cmidrule(l){2-3}
            & ReCoE~\citep{hua2024propagation}
                & QA Accuracy \\
        \cmidrule(l){2-3}
            & \multirow{ 1}{*}{ZsRE + CounterFact$^\dagger$~\cite{Yao2023EditingLL}}
                & Subject-Replace, Reverse-Relation, One-Hop \\
         
    \midrule
        \multirow{ 7}[3]{*}{Locality}
            & \multirow{1}{*}{RippleEdits~\citep{Cohen2023EvaluatingTR}}
                &  Subject Aliasing, Preservation, Relation Specificity \\
        \cmidrule(l){2-3}
            & \multirow{2}{*}{RoundEdit~\citep{li2024unveiling}}
                & Success Score, Distortion ($\downarrow$), Ignore Rate ($\downarrow$), Failure Rate ($\downarrow$), Tied Fact Damage ($\downarrow$)\\
        \cmidrule(l){2-3}
            & \multirow{ 1}{*}{ZsRE + CounterFact$^\dagger$~\cite{Yao2023EditingLL}}
                & Other-Attribution, Distract-Neighbor, Other-Task \\
        \cmidrule(l){2-3}
            & \multirow{2}{*}{CounterFact~\cite{Meng2022LocatingAE}} 
                & Locality, Neighborhood Score, Neighborhood Magnitude \\
        \cmidrule(l){2-3}
            & \multirow{1}{*}{CounterFact+~\cite{hoelscher-obermaier-etal-2023-detecting}}
                & Neighborhood KL Divergence \\
    \bottomrule
    \end{tabular}
    \caption{Performance benchmarks and evaluation metrics addressing generalization/portability and locality issues in knowledge editing methods. Unless specifically indicated by a downward arrow, higher values signify better performance in those evaluation metrics. CounterFact benchmark is proposed by ~\citep{Meng2022LocatingAE}, and CounterFact with $^\dagger$ mark is modified by ~\citep{Yao2023EditingLL} to further examine the proposed metrics.}
    \label{tab:problems_gen_and_loc}
\end{table*}

\subsection{Inability to Logically Inference and Robustly Generalize}
\label{sec:prob-generalization}
When a fact is updated, it is crucial not only to revise the specific piece of knowledge but also to evaluate the impact on the related reasoning chain. Recently the term \textbf{portability} has been proposed in~\citep{Yao2023EditingLL} to evaluate whether an edited fact can be logically inferred within the knowledge chain, and to further assess the robustness of generalization. In their study, they introduce three metrics to evaluate portability: Subject Replace (checking if synonyms of the subject are edited), Reversed Relation (checking if the reversed relation of the target is edited), and One Hop (assessing if modified knowledge is usable for further derivation). Similarly, RippleEdits benchmark and its corresponding Logical Generalization and Compositionality metrics are proposed to examine whether edited knowledge can be inferred in composite relations of facts~\cite{Cohen2023EvaluatingTR}. Additionally, ReCoE benchmark is proposed to assess the propagation of updates in interconnected facts using various reasoning schemes in complex question-answering datasets~\cite{hua2024propagation}. Furthermore, MQuAKE benchmark is introduced to evaluate more complex reasoning and inference ability on multi-hop questions~\cite{zhong2023mquake}.

When multiple logically related facts are edited simultaneously, models may become confused by conflicts between their pre-existing knowledge and the newly edited information. ConflictEdit benchmark is thus proposed to examine different editing methods on conflicted edit facts~\cite{li2024unveiling}. The different benchmarks and corresponding metrics and are arranged systematically in Table~\ref{tab:problems_gen_and_loc}.

\subsection{Unintended Alteration of Non-Target Knowledge}
\label{sec:prob-alteration}
Locality is conventionally assessed using a locality dataset to evaluate the impact of edits on unrelated facts by measuring the Neighborhood Score and Neighborhood Magnitude~\citep[NS \& NM;][]{Meng2022LocatingAE, meng2023memit}. However, current evaluation methods do not adequately capture the post-edit effects on content beyond the locality dataset, which means the edited model could still contain unintended alterations. For example, while the location of the Louvre is successfully modified from Paris to London, the edited model might also output London in an unrelated context or increase the probability of words semantically related to London (e.g., Big Ben) when mentioning the Louvre. Some modified benchmark (CounterFact+) and corresponding metric (Neighborhood KL Divergence)~\citep{hoelscher-obermaier-etal-2023-detecting} is then designed to disclose these previously implicit pitfalls. Another study~\citep{Yao2023EditingLL} extends this exploration to three facets of locality: Other Relations (evaluating the retention of other attributes of the updated subject), Distract Neighborhood (assessing whether model will be swayed by edited cases when they are concatenated before unrelated inputs), and Other Tasks (examining the influence of edits on the performance of other tasks).

Unintended edits to unrelated facts may occur because a single edit can implicitly change the predictive distribution among objects associated with the same (\textit{subject} - \textit{relation}) pair. After multiple consecutive edits, these alterations can accumulate and distort the stored knowledge. To evaluate this condition, the concept of Knowledge Distortion has been introduced by~\citet{li2024unveiling}, which estimates the Jensen–Shannon divergence of the object set distribution before and after editing. This can be further extended to metrics such as the Ignore Rate, measuring how objects other than the target in the object set are neglected after editing, and the Failure Rate, which measures the proportion of instances where over half of the objects in the set are overlooked.

\subsection{Deterioration of General LLM Abilities}
\label{sec:catastrophic}
Current evaluation metrics are primarily limited to scenarios where editing is performed only once or infrequently, prompting some studies to extend evaluations to the outcomes after consecutive edits. A study by \citet{gupta2024model} discovers that post-edit models exhibit susceptibility to both gradual forgetting and catastrophic forgetting in sequential editing scenarios. Notably, their findings indicate that the extent of knowledge forgetting is more pronounced in meta-learning-based methods compared to locate-and-edit methods. Additionally, models with parameters modified successively show a decline in performance across various downstream NLP tasks~\citep{gu2024model}. Furthermore, perplexity is found to increase after consecutive edits across all parameter-modified methods and different LLMs, and is proposed as another metric to indicate model collapse~\citep{yang2024butterfly}. These findings further corroborate that model editing aimed at modifying parameters adversely affects the general capabilities of the original LLMs.

\begin{table*}[ht]
    \centering
    \small
    \begin{tabular}{lcccccccc}
    \toprule
        & \multicolumn{3}{c}{Single Edit} & \multicolumn{5}{c}{Multiple Edit} \\
        \cmidrule(l){2-4} \cmidrule(l){5-9}
        & \multicolumn{3}{c}{One-Hop} & \multicolumn{1}{c}{Multiple-Hop} & \multicolumn{2}{c}{Reverse Conflict} & \multicolumn{2}{c}{Composite Conflict} \\
        \cmidrule(l){2-4} \cmidrule(l){5-5} \cmidrule(l){6-7} \cmidrule(l){8-9}
        Methods & SR & RR & OH & MH & CS & CM & CS & CM \\
        \midrule
        FT & 72.96 & 8.05 & 1.34  & 1.6 & 80.28 & 71.11 & 75.45 & 64.28\\
        MEND & 42.45 & 0.00 & 11.34 & 9.2 & 88.89 & 60.50 & 84.85 & 43.45\\
        ROME & 37.42 & 46.42 & 50.91 & 7.6 & 65.92 & -0.65 & 71.70 & 37.04\\
        MEMIT & 27.73 & 47.67 & 52.74 & 8.1 & 51.40 & -1.60 & 57.15 & -1.50\\
        SERAC & 17.79 & 1.30 & 5.53 & 7.9$^\dagger$ & 50.89$^\dagger$ & -0.02$^\dagger$ & 50.84$^\dagger$ & -0.02$^\dagger$ \\
        IKE & 88.77 & 92.96 & 55.38 & 8.3$^\dagger$  & 58.20$^\dagger$ & -1.00$^\dagger$ & 50.52$^\dagger$ & -0.99$^\dagger$ \\
    \bottomrule
    \end{tabular}
    \caption{Experimental results for portability and generalization. SR: Subject-Replace, RR: Reverse-Relation, OH: One-Hop Accuracy, MH: Multi-hop Accuracy, CS: Conflict score, CM: Conflict magnitude. Higher values indicate better performance for all metrics in this table. Results marked with $\dagger$ are obtained in our own experiments, and other results are taken from previous studies.}
    \label{tab:results_gen_and_loc_1}
\end{table*}

\begin{table*}[ht]
    \centering
    \small
    \begin{tabular}{lccccccccc}
    \toprule
        & \multicolumn{3}{c}{Single Edit} & \multicolumn{5}{c}{Multiple Edit} \\ \cmidrule(l){2-4} \cmidrule(l){5-9}
        Methods & OA & DN & OT & Succ. & D ($\downarrow$) & IR ($\downarrow$) & FR ($\downarrow$) & \\
        \midrule
        FT & 12.88 & 9.48 & 49.56 & 100.0 & 16.12 & 97.48 & 97.32 \\
        MEND & 73.50 & 32.96 & 48.86 & 99.12 & 14.35 & 87.64 & 86.56 \\
        ROME & 78.94 & 50.35 & 52.12 & 99.80 & 13.95 & 78.98 & 77.60 \\
        MEMIT & 86.78 & 60.47 & 74.62 & 99.72 & 13.50 & 72.03 & 70.44 \\
        SERAC & 99.50 & 39.18 & 74.84 & 50.14$^\dagger$ & 3.78$^\dagger$ & 99.62$^\dagger$ & 99.64$^\dagger$ \\
        IKE & 84.13 & 66.04 & 75.33 & 100.0$^\dagger$ & 13.43$^\dagger$ & 73.53$^\dagger$ & 73.00$^\dagger$ \\
    \bottomrule
    \end{tabular}
    \caption{Experimental results for locality. OA: Other-Attribution, DN: Distract-Neighbor, OT: Other-Task, Succ.: Success rate, D: Distortion, IR: Ignore rate, FR: Failure rate. Unless specifically indicated by a downward arrow, higher values signify better performance in those evaluation metrics. Results marked with $\dagger$ are obtained in our own experiments, and other results are taken from previous studies.}
    \label{tab:results_gen_and_loc_2}
\end{table*}

\section{Experiments}
\label{sec:experiments}
The experiments are done to evaluate robust generalization and locality (Section ~\ref{sec: gen and loc} as well as deterioration of general LLM abilities (Section ~\ref{sec:gen ability} across different editing methods.

\subsection{Experimental Setup}
Given the variety of benchmarks addressing different challenges in knowledge editing, systematically comparing model performance becomes difficult. To address this, we select the most widely used datasets for each category of pitfalls, ensuring a fair and transparent comparison.
\subsubsection{Robust Generalization and Locality}
\label{sec: gen and loc}
We use GPT-J \citep{gpt-j} as the baseline model for editing and implement six distinct editing methodologies to assess robust generalization and locality: MEND (meta-learning), ROME and MEMIT (locate-and-edit), SERAC (external memory), and IKE (prompting).

Given the overlap in benchmarks for robust generalization and locality, we select a subset for our experiments. The evaluation is divided into two settings: \emph{single edit}, where only one fact in a reasoning chain is modified, and \emph{multiple edits}, where several logically connected facts in the chain are altered simultaneously. A detailed description is provided in the Appendix~\ref{appendix:metrics}). Single edit metrics include Subject-Replace, Reverse-Replace, and One-Hop reasoning~\citep{Yao2023EditingLL}. Multiple edit metrics include multi-hop editing accuracy~\citep{zhong2023mquake}, and Conflict Score and Conflict Magnitude for Reverse Conflict and Composite Conflict respectively~\citep{li2024unveiling}. For locality, single edit metrics include Other-Attribution, Distract-Neighbor, and Other-Task~\citep{Yao2023EditingLL}, while multiple edit metrics encompass Success Rate, Distortion, Ignore Rate, and Failure Rate~\citep{li2024unveiling}.

\begin{figure*}[ht]
    \centering
    \begin{subfigure}{0.45\linewidth}
        \includegraphics[width=\linewidth]{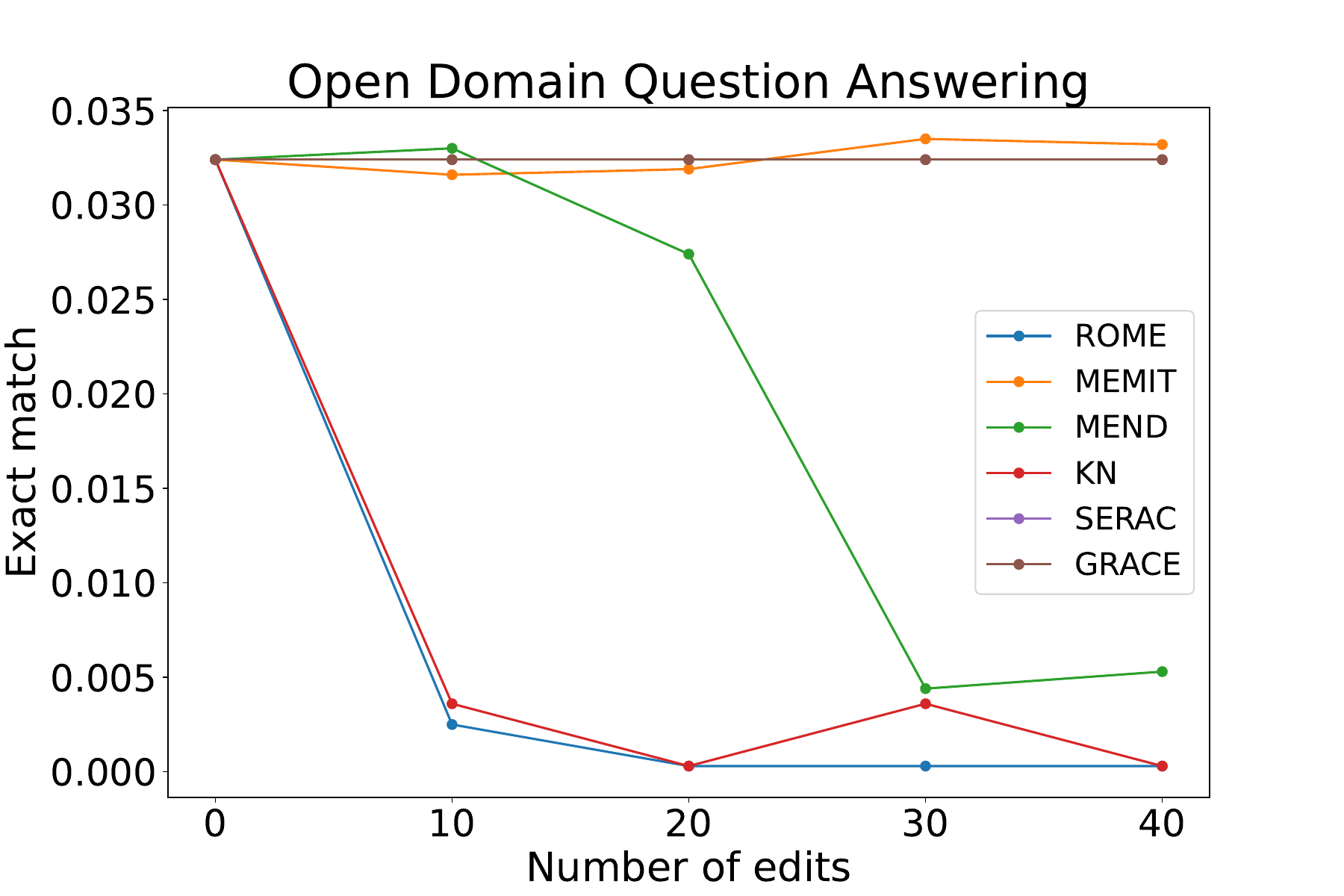}
        \caption{Open Domain Question Answering}
        \label{fig:opendomain}
    \end{subfigure}
    \begin{subfigure}{0.45\linewidth}
        \includegraphics[width=\linewidth]{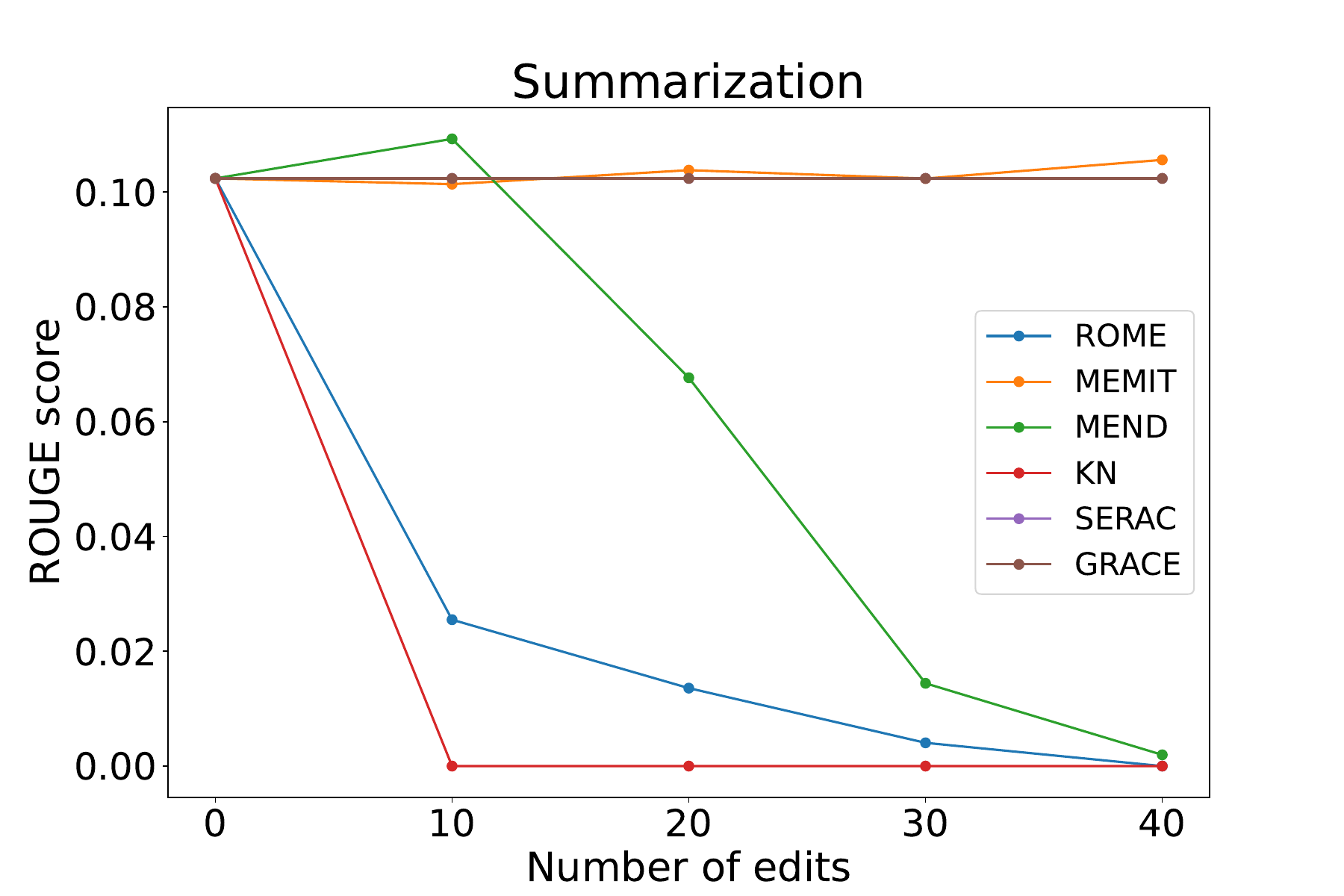}
        \caption{Summarization}
        \label{fig:summarization}
    \end{subfigure}
    \begin{subfigure}{0.45\linewidth}
        \includegraphics[width=\linewidth]{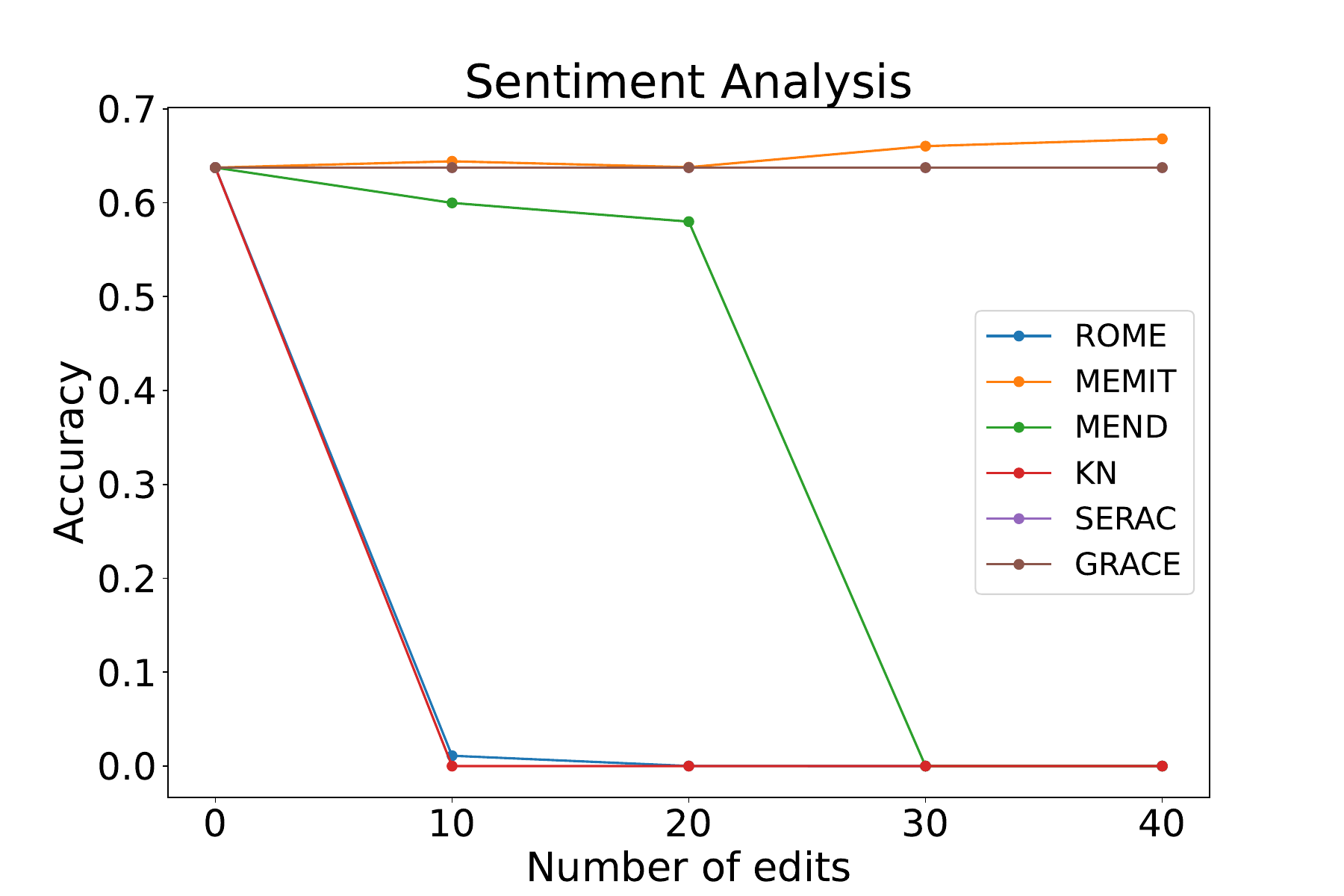}
        \caption{Sentiment Analysis}
        \label{fig:sentiment}
    \end{subfigure}
    \begin{subfigure}{0.45\linewidth}
        \includegraphics[width=\linewidth]{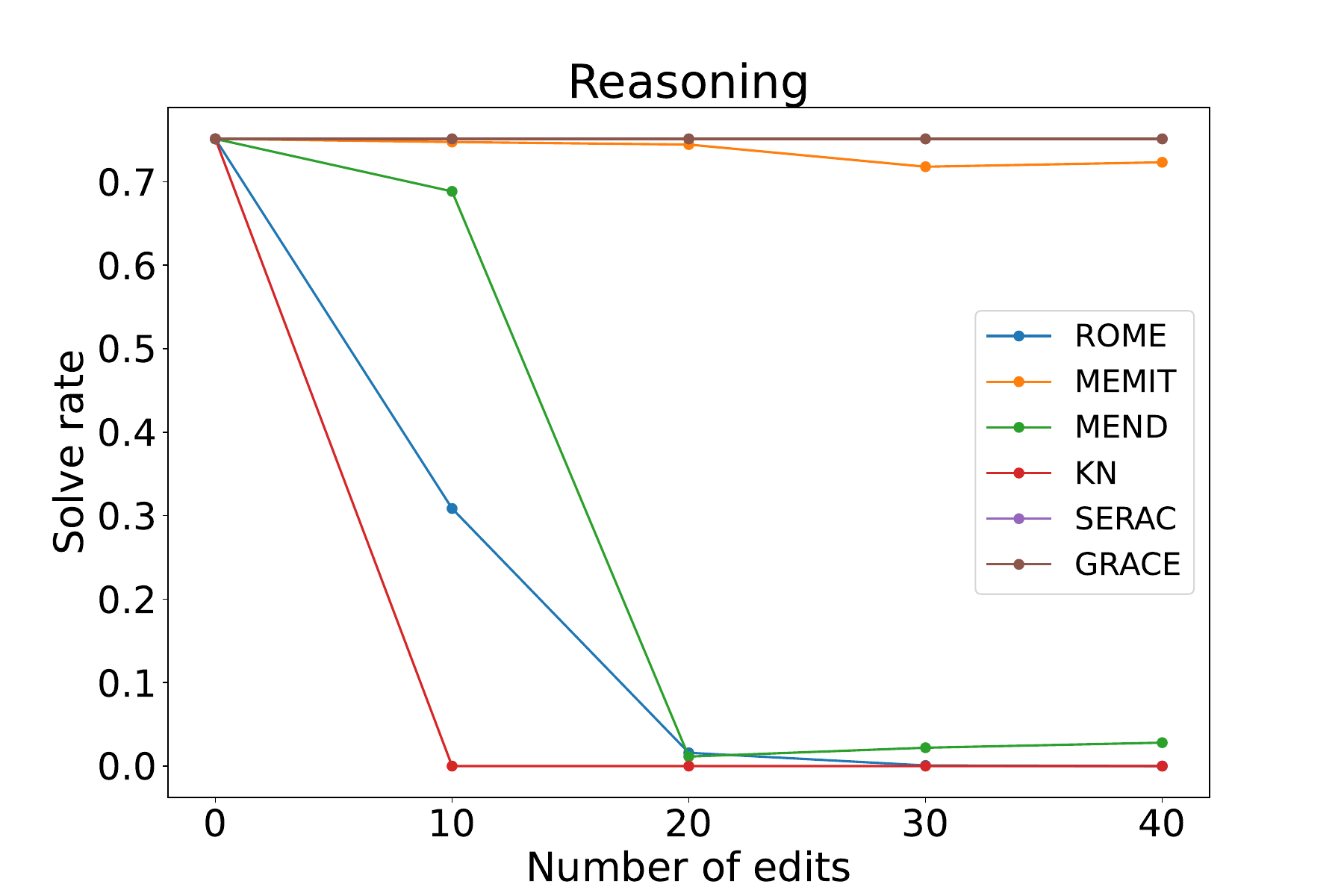}
        \caption{Reasoning}
        \label{fig:reasoning}
    \end{subfigure}
    \caption{The experimental results for the deterioration of general abilities were obtained by editing GPT-J with various editing algorithms, including ROME, MEMIT, MEND, KN, SERAC, and GRACE, each applied 10 to 40 times. The edited models were subsequently evaluated on four downstream tasks, including open-domain question answering, sentiment analysis, summarization, and reasoning. The results for SERAC and GRACE are overlapping.}
    \label{fig:catastrophic forgetting}
\end{figure*}

\subsubsection{Deterioration of General LLM Abilities}
\label{sec:gen ability} 
Following the settings of \citep{gu2024model}, we assess deterioration of general LLM abilities post-editing using six methodologies: ROME, MEMIT, SERAC, MEND, KN, and GRACE. We evaluate general abilities across four NLP downstream tasks: open-domain question answering, sentiment analysis, reasoning, and summarization. These tasks are assessed after 10 to 40 edits on the Zero-Shot Relation Extraction (ZsRE) dataset\citep{levy-etal-2017-zero}, comparing the results against pre-editing benchmarks. More details on the selected downstream tasks are in Appendix~\ref{appendix:experiment}.

\subsection{Experimental Results and Discussion}
In general, current editing methodologies show suboptimal performance in both robust generalization and locality. Regarding robust generalization (Table~\ref{tab:results_gen_and_loc_1}), IKE, which leverages prompt demonstrations, excels in single edit but declines with multiple edits. This suggests that prompt demonstrations may become confused when editing multiple logically related facts. Conversely, fine-tuning and meta-learning-based methods are less susceptible to confusion after editing multiple related facts.

Regarding locality (Table~\ref{tab:results_gen_and_loc_2}), IKE maintains stable performance across metrics in single edit settings. Parameter-modifying methods excel in Other Attribution but decline in other metrics, except MEMIT, which remains stable across all metrics. In multiple edit scenarios, all methods except SERAC show similar performance. In the multiple edit scenario, all methods except SERAC exhibit relatively similar performance. SERAC displays low edit success rate and distortion rate, suggesting its scope classifier does not adopt most edits in this scenario. This may be attributed to its weakness in recovering edited facts, which is crucial in this metric setting.

In terms of general LLM abilities (Figure~\ref{fig:catastrophic forgetting}), the number of edits affects methods differently. Meta-learning methods like MEND degrade significantly after 10-20 edits. Locate-and-edit methods such as ROME and KN degrade after 10 edits, while MEMIT remains stable after 40 edits. This disparity can be attributed to MEMIT's strategy of adjusting parameters across multiple layers, as opposed to ROME's single-layer edits and KN's approach of modifying a few neurons. This distribution of parameter modifications across layers may help mitigate deterioration.

GRACE, which stores edited facts with additional parameters, shows no performance change in downstream tasks after edits. One possible explanation is that the edits are conducted on the ZsRE dataset, which is distinct from the requirements of downstream tasks, leading to the stored facts not being retrieved during inference. Similarly, SERAC, utilizing external memory for edited facts, preserves general NLP abilities post-editing. This preservation stems from SERAC being trained once before editing begins, solely performing inference during editing, thereby preventing changes in the model's output, even after multiple edits.

Overall, parameter-modifying methods degrade downstream task performance by altering pre-trained LLM parameters. In contrast, parameter-preserving methods maintain the original parameters, resulting in stable downstream task performance even after multiple edits.

\section{Future Prospects}
\label{sec:limit_chall_future}
\subsection{Leveraging Information Retrieval and External Memory}
Research shows that using external knowledge bases, rather than relying solely on internal knowledge, benefits LLMs by guiding content generation based on predefined facts. External knowledge sources, such as text corpora, structured tables, or key-value databases, can be utilized either to finetune LLMs for improved information retrieval or to employ prompting techniques for querying these sources. These approaches separate factual knowledge from inference process, thus preserves the original model parameters and minimizes post-editing damage. Moreover, they ensure that generated content aligns with predefined knowledge bases, thereby enhancing accountability and accuracy.


\subsection{Improving Understandings of LLMs' Internal Knowledge Structures}
While identifying where factual knowledge is stored in LLMs has been extensively explored~\citep{Meng2022LocatingAE, meng2023memit, dai-etal-2022-knowledge, hernandez2024linearity, geva-etal-2021-transformer}, the correlation between knowledge location and editing success remains low~\citep{hase2023does}. Additionally, despite evidence suggesting a strong connection between factual knowledge and the feed-forward network layers~\citep{Meng2022LocatingAE, geva-etal-2021-transformer, geva-etal-2022-transformer}, recent findings indicate that updates to multi-head self-attention layers also improve outcomes~\citep{Li2023PMETPM}. This suggests that locating fact storage alone doesn't fully explain knowledge structures in LLMs. Further research is needed to understand how knowledge locations interact with model predictions in order to enhance LLM interpretability and controllability.

Preserving LLMs' general capabilities is also crucial for model editing, as discussed in Section~\ref{sec:catastrophic}. Recent breakthroughs in identifying regions within models that correlate with general linguistic abilities have opened up a direction for future research in model editing~\citep{zhang2024unveiling}. By making targeted modifications, we can potentially prevent the deterioration of general abilities and improve the specificity and effectiveness of model editing methods.

\subsection{Improving Robustness of Knowledge Editing}
Even after achieving fair scores on the existing metrics, models may revert to pre-edit versions or provide ambiguous answers if the altered knowledge is conflicted with inherited concepts. Experiments show that more popular knowledge is easier for modified models to revert to~\citep{ma2024is}, indicating the lack of robustness in current editing strategies. A deeper understanding of how LLMs store and process interconnected knowledge entities is crucial for more robust editing and warrants future research.

\section{Conclusion}
Although model editing techniques appear promising for cost-effectively updating knowledge, they still have significant pitfalls. Current editing methods often struggle with making logical inferences based on the edited facts, introducing unintended alterations of non-target knowledge and deterioration in model performance, particularly with parameter-modified methods. By harnessing information retrieval techniques and delving into how models store and process knowledge, deviations in model abilities can be mitigated, and the controllability of edited facts can be enhanced, ultimately leading to greater robustness. We hope our work illuminates potential directions for future improvements in knowledge editing. 

\section*{Limitations}
The field of knowledge editing is advancing at an impressive pace, with numerous innovations in editing methodologies and evaluation metrics being proposed. Despite our efforts to collect and organize previous work, some contributions may not be included in this paper. However, we will continue to monitor the latest developments in this field and update our GitHub repository with recent related works.

\section*{Acknowledgments}
We thank the reviewers for their insightful comments.
This work was financially supported by the National Science and Technology Council (NSTC) in Taiwan, under Grants 111-2222-E-002-013-MY3 and 112-2223-E002-012-MY5. 
We thank to National Center for High-performance Computing (NCHC) of National Applied Research Laboratories (NARLabs) in Taiwan for providing computational and storage resources.
\bibliography{acl_latex}

\newpage
\appendix

\section{Detailed Explanation of Evaluation Metrics and Examples}

\label{appendix:metrics}

\subsection{Portability / Generalization}

\paragraph{Single Edit} In the single edit scenario, we modify only one fact in the logical chain with each edit. Let $Z_e = \{ (x_e, y_e) \ | \ f_{\theta_0}(x_e) \neq y_e \}$ be the set where only a single fact is edited in each logical chain. Single edit is conducted as:
\begin{equation}
    f_{\theta_e}(x_e) = y_e, \forall (x_e, y_e) \in Z_e
\end{equation}

This part consists of:
\begin{itemize}
\item One-Hop: This setting focuses on evaluating the impact of a single edit on direct, one-hop reasoning tasks.
\end{itemize}

For one-hop evaluations, we adopt the methods proposed by~\citep{Yao2023EditingLL}. These include:

\begin{itemize}
\item \textbf{Subject Replace:} This metric tests the model's generalization ability by replacing the subject in the question with an alias or synonym, assessing if the edited attribute is generalized to other descriptions of the same subject.
\item \textbf{Reversed Relation:} This metric evaluates the model's capability to handle reversed relations by filtering for suitable relations (e.g., one-to-one relation) and asking the reverse question to check if the target entity is also updated.
\item \textbf{One-Hop Test:} This metric assesses the edited language model's performance on downstream tasks that require one-hop reasoning.
\end{itemize}

\paragraph{Multiple Edits}
In the multiple edits scenario, we evaluate the model's performance after applying several logically related edits. Let $Z_e = \{ (x_{ei}, y_{ei}) \ | \ f_{\theta_0}(x_{ei}) \neq y_{ei} \}$ represent a set of logically related facts within a reasoning chain intended to be modified. Multiple edits are performed by altering several facts within this chain:
\begin{equation}
    f_{\theta_e}(x_{ei}) = y_{ei}, \forall (x_{ei}, y_{ei}) \in Z_e
\end{equation}

This part consists of:
\begin{itemize}
    \item Multi-Hop editing: Evaluate whether the model can infer edited knowledge in multi-hop questions.
    \item Conflict editing: Assess how the model handles multiple conflicting edits.
\end{itemize}

In the multi-hop setting, we assess the model's performance on multi-hop questions using the evaluation methods proposed by~\citep{zhong2023mquake}, which include:

\begin{itemize}
\item \textbf{Edit-wise Success Rate (EW):} This metric measures how many facts can be successfully recalled from the edited language model.
\begin{equation}
    \mathrm{EW} = \mathbbm{1}\{f^*(s) =  o^*\}
\end{equation}
where $f^*$ is the model after editing, $s$ refers to the edited subject, and $o$ refers to target object.
\item \textbf{Instance-wise Accuracy (IW):} This metric tests how many multi-hop instances the model can recall all the individual single-hop facts. This metric is crucial for multi-hop performance, as the model must encode each fact to answer the multi-hop question. 
\begin{equation}
    \mathrm{IW} = \mathbbm{1}\{\bigwedge_{(s, r, o^*) \in C^*} [f^*(s) = o^*]\}
\end{equation}
where $C^* = \langle(s_1, r_1, o_1), \dots, (s_n, r_n, o_n) \rangle$ is the chain of facts of a multi-hop question. In this chain, the object of the $i^{\mathrm{th}}$ fact is the subject of the next fact. (i.e., $o_i = s_{i+1}$)
\item \textbf{Multi-hop Accuracy (MH):} This metric assesses the accuracy of the original and edited language models on multi-hop questions. In the MQuAKE dataset~\citep{zhong2023mquake}, there are three generated multi-hop questions for each instance. If any of the three questions is correctly answered by the model, we consider it accurate.
\begin{equation}
    \mathrm{MH} = \mathbbm{1}\{ \bigvee_{q \in Q} f^*(q) = a^*\}
\end{equation}
where $Q$ is a set of similar multi-hop questions with the same answer $a^*$.
\end{itemize}

As for Conflict editing, we use the setting and evaluation methods from~\citep{li2024unveiling}. The settings consist of:

\begin{itemize}
\item \textbf{Reverse Conflict:} This setting introduces conflicts by editing facts with reverse relations. For example: 
\\\textbf{edit 1:} (\(s_{1}, r_{1}, o_{1}\)→\(o_{2}\))
\\\textit{Hamlet was written by Shakespeare} → \textit{Agatha Christie}. 
\\\textbf{edit 2:} (\(o_{2}, r_{2}, s_{1}\)→\(s_{2}\))
\\\textit{The notable work of Agatha Christie is Hamlet} → \textit{Odyssey}
\\the updated knowledge then could be represented as:
\[
\left\{
\begin{array}{l}
k_o = (s_1, r_1, o_2) \\
k_n = (s_2, r_1, o_2)
\end{array}
\right.
\]
where $k_o$ refers to old knowledge, and $k_n$ refers to new knowledge.
\item \textbf{Composite Conflict:} This explores more complex situations where the edits are associated with a fact that is not influenced by the editing (\textbf{tied fact}). For example:
\\\textbf{edit 1:} (\(s_{1}, r_{1}, o_{1}\)→\(o_{2}\))
\\\textit{Hamlet was written in English} → \textit{French}
\\\textbf{edit 2:} (\(s_{2}, r_{2}, o_{2}\)→\(o_{3}\))
\\\textit{Shakespeare wrote in French} → \textit{German} 
\\\textbf{tied fact:} (\(s_{1}, r, s_{2}\))
\\\textit{The notable work of Shakespeare is Hamlet}
\\where \( r \land r_1 \to r_2 \) is a logical rule. The updated knowledge then could be represented as:
\[
\left\{
\begin{array}{l}
k_f = (s_1, r, s_2) \\
k_0 = (s_1, r_1, o_2) \\
k_n = (s_1, r_1, o_3)
\end{array}
\right.
\]
where $k_f$ refers to a tied fact.
\end{itemize}

The evaluation methods include:

\begin{itemize}
\item \textbf{Conflict Score (CS):} Measures how well a knowledge editing method handles knowledge conflicts by calculating the ratio that the new fact is more probable than the old fact after knowledge editing.
    \begin{equation}
        \mathrm{CS = } \mathbbm{1}\{p_{f_\theta'}(k_n) > p_{f_\theta'}(k_o)\}
    \end{equation}
\item \textbf{Conflict Magnitude (CM):} Estimates the decrease in probability of the old fact after editing.
    \begin{equation}
        \mathrm{CM = } \frac{p_{f_{\theta^m}}(k_{o}) - p_{f_{\theta'}}(k_{o})}{p_{f_{\theta^m}}(k_{o})}
    \end{equation}
\( \theta^m \) is the intermediate model parameters after \textit{edit 1}.
\end{itemize}

\subsection{Locality}

\paragraph{Single Edit}
In the single edit scenario for locality, we adopt the methods proposed by \citep{Yao2023EditingLL}, including:
\begin{itemize}
    \item \textbf{Other Attribution (OA)}:
    The modified \textbf{ZsRE} and \textbf{CounterFact} datasets are applied to test whether the non-target attributes of the edited subjects remained the same. For example, if we reset \textit{Lionel Messi} as a basketball player, his nationality should stay the same. 
    \item \textbf{Distract Neighbor (DN)}:
    Previous studies indicate that if edit cases are concatenated with unrelated context, the model tends to output content related to the edit cases. For example, if the original prompt is "Windows 11 is a product of \_\_", an edit case is added in front and be "Windows 11 is a product of Google. Office 365, developed by \_\_". It testifies whether the model prediction would be "distracted" by the edit case.
    \item \textbf{Other Task (OT)}
    The edited model is tested on the multiple-choice QA task \textbf{Physical Interaction QA} (PIQA, \citet{Bisk_Zellers_Lebras_Gao_Choi_2020}) and the performance is evaluated by accuracy. 
\end{itemize}

\paragraph{Multiple Edits}
We also test the model's locality in the multiple edits scenario by adopting the methods and evaluations from~\citep{li2024unveiling}. The settings consist of:

\begin{itemize}
    \item \textbf{Round Edit:} This edits the knowledge triplet back-and-forth, for example:
    \\\textbf{edit 1:} (\(s, r, o_{1}\)→\(o*\))
    \\\textbf{edit 2:} (\(s, r, o*\)→\(o_{1}\))
    
    where $o^*$ is an intermediate object.
\end{itemize}
The evaluation metrics include: 
\begin{itemize}
    \item \textbf{Distortion (D)~\citep{li2024unveiling}:}
        \begin{equation}
            D = JS\left( p_{f_\theta}(\text{Obj} \mid (s, r)), p_{f_{\theta'}}(\text{Obj} \mid (s, r)) \right)
        \end{equation}
    estimates the JS divergence of the objects distribution before and after edit.
    
    \item \textbf{Ignore Rate (IR)~\citep{li2024unveiling}:}
        \begin{align}
            \begin{split}
                \mathrm{IR =}\frac{1}{\left|\mathrm{Obj}\right| - 1} \sum_{\substack{o \in \mathrm{Obj}\setminus\{o1\}}}
                &\mathbbm{1}\{p_{f_\theta}(o \mid (s, r)) > \\ &p_{f_\theta'}(o \mid (s, r))\}
            \end{split}
        \end{align}
    measures the extent to which objects in Obj set (excluding the target object $o_1$) are disregarded or overlooked after the process of knowledge editing.
    
    \item \textbf{Failure Rate (FR)~\citep{li2024unveiling}:}
        \begin{equation}
            \mathrm{FR =}\mathbbm{1}\{ \mathrm{IR} > 0.5 \}
        \end{equation}
    calculates the rate when Ignore Rate > 0.5
    \item \textbf{Tied Fact Damage (TDF)~\citep{li2024unveiling}:}
    \begin{equation}
        \mathrm{TFD = } \frac{p_{f_{\theta^m}}(k_{f}) - p_{f_{\theta'}}(k_{f})}{p_{f_{\theta^m}}(k_{f})}
    \end{equation}
    \( k_{f} \) denotes the tied facts and \( \theta^m \) is the intermediate model parameters after \textit{edit 1}.
       
\end{itemize}

\paragraph{Other Locality Metrics}
\begin{itemize}
    \item \textbf{Neighborhood KL Divergence~\citep{hoelscher-obermaier-etal-2023-detecting}:}
    \begin{equation}
    \mathrm{NKL}\stackrel{\mathrm{def}}{=} \sum_{w \in W}\log\left(\frac{P(w)}{P^*(w)}\right)
    \end{equation}
    \item \textbf{Neighborhood Score (NS)~\citep{Meng2022LocatingAE}:} collect a set of "neighborhood" subjects and evaluate the success fraction for $\mathbb{P}\left[o^{c}\right]>\mathbb{P}\left[o^{*}\right]$, while the $o^c$ denotes the correct facts and $o^*$ denotes the false facts. 
    
    \item \textbf{Neighborhood Magnitude (NM)~\citep{Meng2022LocatingAE}:} the differences of $\mathbb{P}\left[o^{c}\right]$ and $\mathbb{P}\left[o^{*}\right]$ for the "neighborhood" subjects.
   
\end{itemize}

\section{Detailed Experimental Details of the Deterioration of General LLM Abilities}
\label{appendix:experiment}
We follow the settings of ~\citep{gu2024model} for this part of experiments. Different evaluation metrics were applied for each downstream task: Exact Match for open-domain question answering on the Natural Question dataset ~\citep{kwiatkowski-etal-2019-natural}, accuracy for sentiment analysis on the SST2 dataset ~\citep{socher-etal-2013-recursive}, solve rate for reasoning on the GSM8K dataset ~\citep{Cobbe2021TrainingVT}, and ROUGE score for summarization on the SAMSum dataset ~\citep{gliwa-etal-2019-samsum}.

\end{document}